\documentclass[10pt,twocolumn,letterpaper]{article}

\usepackage{cvpr}              
\usepackage{booktabs}  
\usepackage{stfloats}
\usepackage{placeins}

\usepackage{multirow}
\usepackage{comment}
\usepackage{amssymb}
\usepackage{pifont}
\newcommand{\cmark}{\ding{51}}%
\newcommand{\xmark}{\ding{55}}%
\usepackage[accsupp]{axessibility}
\definecolor{cvprblue}{rgb}{0.21,0.49,0.74}
\usepackage[pagebackref,breaklinks,colorlinks,allcolors=cvprblue]{hyperref}
\def\paperID{6} 
\def\confName{CVPR}
\def\confYear{2025}

\title{Slot Attention-based Feature Filtering for Few-Shot Learning}

\author{Javier Ródenas\\
AIBA, Departament de \\
 Matemàtiques \&  
Informàtica  \\
Universitat de Barcelona\\
Barcelona, Spain\\
{\tt\small jrodencu33@alumnes.ub.edu}
\and
Eduardo Aguilar\\
Departamento de Ingeniería de \\
Sistemas y Computación \\
Universidad Católica del Norte\\
Antofagasta, Chile\\
AIBA, Universitat de Barcelona\\
Barcelona, Spain\\
{\tt\small eduardo.aguilar@ub.edu}
\and
Petia Radeva\\
 AIBA, Departament de \\
 Matemàtiques \&  
Informàtica  
\\ Institute of Neuroscience, 
\\
Universitat de Barcelona\\
Barcelona, Spain\\
{\tt\small petia.ivanova@ub.edu}
}

\newcommand{\museNo}{01070\-421}
\newcommand{\DFVolNo}{PDC\-2022-133642-I00}

\begin{document}
\maketitle

\begin{abstract}
Irrelevant features can significantly degrade few-shot learning performance. This problem is used to match queries and support images based on meaningful similarities despite the limited data. However, in this process, non-relevant features such as background elements can easily lead to confusion and misclassification. To address this issue, we propose Slot Attention-based Feature Filtering for Few-Shot Learning (SAFF) that leverages slot attention mechanisms to discriminate and filter weak features, thereby improving few-shot classification performance. The key innovation of SAFF lies in its integration of slot attention with patch embeddings, unifying class-aware slots into a single attention mechanism to filter irrelevant features effectively. We introduce a similarity matrix that computes across support and query images to quantify the relevance of filtered embeddings for classification. Through experiments, we demonstrate that Slot Attention performs better than other attention mechanisms, capturing discriminative features while reducing irrelevant information. We validate our approach through extensive experiments on few-shot learning benchmarks: CIFAR-FS, FC100, miniImageNet and tieredImageNet, outperforming several state-of-the-art methods.
\end{abstract}

\section{Introduction}
In recent years, the machine learning research community has invested significant effort into developing new perspectives to simplify and generalize the classification of new categories in datasets. Traditionally, adding a new category to a classification model required a large number of images which can be resource and time-consuming. However, the new paradigm of  Few-Shot Learning (FSL) opens the door to finding a solution to this issue using a few images to classify new categories. 

Few-shot learning emerged as an innovative approach to generalize and use the acquired knowledge from base categories reducing the number of labeled samples needed. FSL provides a more flexible and scalable solution than traditional approaches. Within the number of labeled samples required, we find different approaches such as Zero-Shot Learning \cite{Stojni?_2024_CVPR,mirza2023lafter} where the model learns based on known classes without having any example from the new class, One-Shot Learning \cite{Chen_2019_CVPR, yu2024learningshot} where the model just has one example per each new class or Few-Shot learning \cite{Bar_2024_CVPR, semantic} where models are trained with a small number of examples per class. The approach of learning from limited examples, means rapid adaptation to novel tasks and efficient learning in domains where data acquisition requires big efforts.

Despite its potential, FSL presents big challenges. The task of accurately classifying new categories based on a limited number of examples remains difficult, requiring concentrating efforts on creating sophisticated models and focusing on relevant features. The main difficulty lies in extracting the key features and getting the best representation for every category.

On the other hand, FSL is easily prone to overfitting \cite{overfitting}. Due to the data scarcity, models can easily memorize the training data without learning generalizable features. To address this problem, researchers have proposed different methodologies. For instance, MAML \cite{MAML} introduced a method to train a model on a variety of learning tasks, so that it can solve new learning tasks using only a small number of training samples without overfitting.

Another significant challenge is the issue of feature relevance. In traditional machine learning with large datasets, models can learn to ignore irrelevant features through extensive training. However, in FSL the limited number of examples makes it crucial to focus on the most discriminative features from the outset. This is particularly important in image classification tasks, where background elements or other non-relevant features can lead to misclassification. To address these challenges, researchers have been working on different approaches. Metric-based methods such as the very recent study \cite{lai2024clusteredpatchelementconnectionfewshot}, employ a cosine similarity metric mechanism between query and support images to filter the most important features, mitigating the impact of irrelevant features before matching. Also, attention mechanisms have been employed to focus on relevant parts of the image, as seen in the work \cite{wang2023focus}.

\begin{figure}[t!]
\centering
\includegraphics[width=1\columnwidth, height=0.23\textheight]{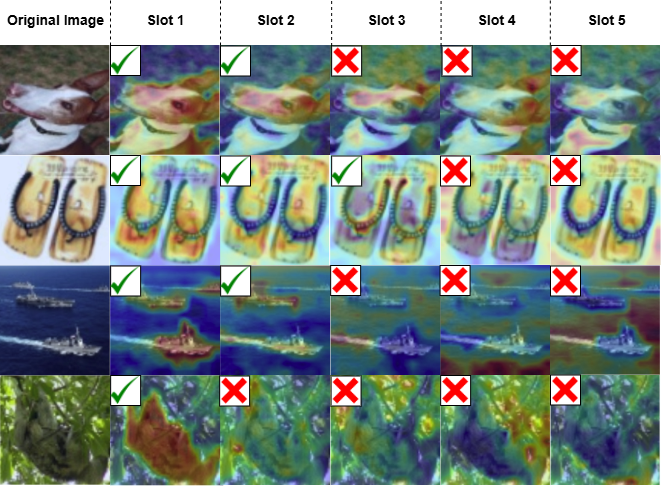}
\caption{Visualization of the attention assigned to each slot. Green check marks indicate those that passed the filter, while red crosses denote those that were discarded.}
\label{fig:refined_masks}
\end{figure}

Our proposed method, Slot Attention-based Feature Filtering (SAFF), addresses the problem of feature relevance in a novel way. By integrating slot attention into the FSL pipeline, SAFF aims to focus on the most discriminative features and improve classification performance. We adapt the slot attention mechanism, originally proposed by Locatello \cite{locatello2020objectcentriclearningslotattention} to the FSL context, using it to identify and focus on the most relevant features for classification. SAFF decomposes the embedding extracted from the model into slots to get attentions from the different embeddings where every slot tries to {\em fight} for the attention of a group of features. By comparing these slots with the class token, we can identify which are best aligned with the class token. This comparison acts as a filtering mechanism that isolates the slots contributing most effectively to the class representation. The selected slots are combined to create a refined attention map, which is then integrated back into the feature set in a weighted manner. This approach ensures that the final feature representation is enriched with information that is most relevant to the target class. In addition, it allows us to filter out irrelevant background elements and focus on the key features that define each category (see  Fig.~\ref{fig:refined_masks}). We apply our method to four public datasets, namely: CIFAR-FS  \cite{CIFAR-FS}, FC100 \cite{FC100}, miniImageNet \cite{MatchingNetwork} and tieredImageNet \cite{tieredImageNet} with very encouraging results.

Our contributions are summarized as follows:

\begin{itemize}
 \item We propose SAFF, a novel method that filters irrelevant features using the slot attention mechanism that effectively discriminates between relevant and irrelevant features in both support and query images.
 \item We introduce combined attention for filtering patch embeddings made from the slots that most represent the class.
 \item We demonstrate the effectiveness of SAFF through extensive experimentation on CIFAR-FS, FC100, miniImageNet, tieredImageNet and improving the state-of-the-art methods.
 \item We provide comprehensive visualization analysis showing that SAFF successfully filters out non-relevant class-aware features.
\end{itemize}

The rest of the paper is structured as follows. In Section \ref{sec:related_work}, we present the related work to our method. Section \ref{sec:methodology} defines our method SAFF and its different steps. In Section \ref{sec:experiments}, we show the results on four public datasets and comparison to the state-of-the-art. Finally, conclusions are given in Section \ref{sec:conclusions}, where we share the final thoughts, limitations and future works.

\section{Related Work}
\label{sec:related_work}

In this section, we briefly discuss the most recent and relevant work on FSL, feature filtering techniques and attention mechanisms. 

\subsection{Few-shot Learning}

Few-shot learning (FSL) has gained significant attention in recent years as a critical area of research in computer vision. Existing approaches can be broadly categorized into optimization-based methods, metric-based methods, and transfer learning-based methods.

\textbf{Optimization-based methods}, aims to enhance the learning process by optimizing model parameters for better generalization to novel tasks. A prominent example is MAML \cite{MAML}, which focuses on learning an initial set of parameters that can be quickly adapted using a few samples from a new task. Similarly, Reptile \cite{Reptile} simplifies meta-learning by reducing the computational cost associated with parameter updates across different tasks.

\textbf{Metric-based methods} focus on obtaining highly representative embeddings for query and support images, ensuring that samples of the same class cluster closely in the feature space. These methods first learn a discriminative feature representation and then apply a similarity metric to identify the most relevant support sample for a given query \cite{NIPS2017_cb8da676, Hao_2023_ICCV, MatchingNetwork}.

\textbf{Transfer learning-based methods} leverage large-scale datasets to extract generalizable features from a related domain. These models serve as feature extractors, enabling the transfer of learned representations to new tasks with minimal data. Fine-tuning such models accelerates learning and improves performance due to the prior knowledge embedded in the features. For instance, MTL \cite{MTL} utilizes a vast collection of similar few-shot tasks to effectively adapt a pre-trained model to a new task with only a few labeled examples.

\subsection{Feature Filtering}
In FSL scenarios, where data is inherently scarce, feature filtering is crucial for identifying the most relevant information and improving representation learning, However, this process presents several challenges. First, the importance of features learned from one task may not generalize well to others. Second, the limited number of examples makes it difficult to reliably distinguish between relevant and irrelevant features \cite{aligment, taskguided}. Consequently, the combination of data scarcity and noisy features, such as background elements, increases the risk of overfitting and misclassification \cite{graph, semantic}. 

\begin{figure*}[t!]
\centering
\includegraphics[width=0.8\textwidth, height=0.27\textheight]{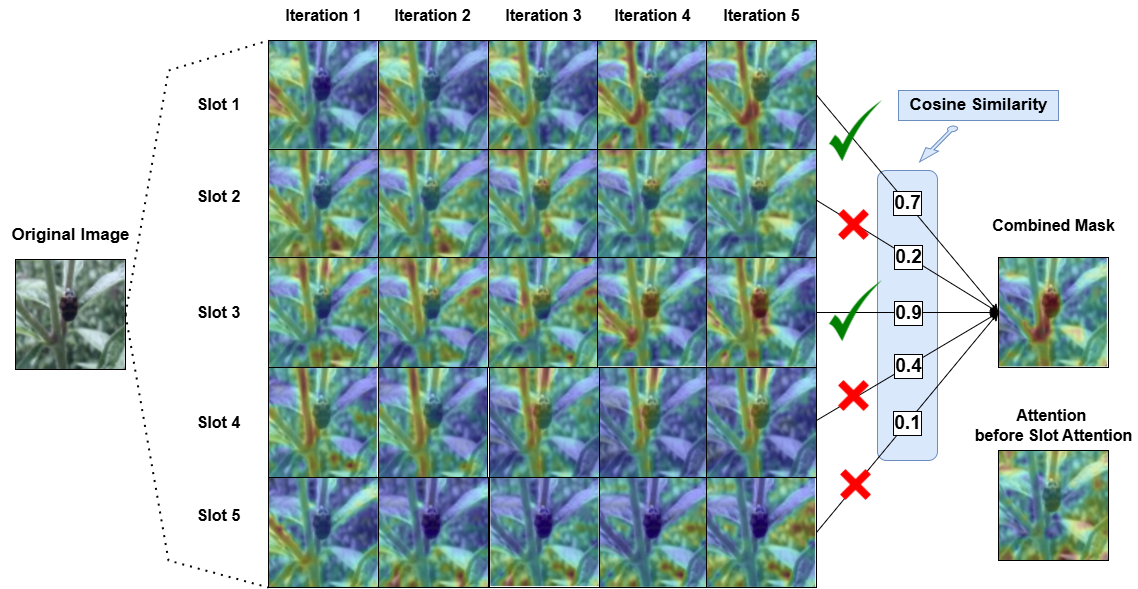}
\caption{First, the attention maps for each of the 5 slots, focus on different regions of the image. Then, iterative refinement enhances relevant features across multiple iterations. Finally, a combined mask is generated from slots that exceed a similarity threshold with the class token, effectively highlighting class-specific areas while under-weighting non-relevant regions. }
\label{fig:iterations}
\end{figure*}

Recent advancements in feature filtering strategies have mitigated these issues. For instance, \cite{Hilbert} proposed a method based on the Hilbert-Schmidt Independence Criterion to enhance feature selection in FSL settings. Additionally, attention mechanisms have proven highly effective in refining feature representations, particularly in low-data scenarios \cite{dual, rectification, SaberNet}. These mechanisms allow the model to focus on both global and local features, effectively filtering out irrelevant information and ensuring that learned representations remain robust and task-specific.

\subsection{Attention Mechanisms}

One way to enable models to focus on the most relevant features is by using attention mechanisms playing an important role in changing the attention depending on the proposed task during the learning process. In FSL, attention mechanisms help in several key aspects. First, attention mechanisms allow models to adapt the focus depending on the task reducing the irrelevant features and removing the noise that can lead to overfitting or misclassification \cite{Lee_2024_CVPR, Fan_2024_CVPR}. Second, these mechanisms improve the generalization across the task enabling more meaningful relationships between support and query samples. Slot attention has emerged as a promising attention mechanism in which each slot can specialize in capturing distinct features which is very valuable for FSL due to the iterations for progressive refinement of representations \cite{SCOUTER}.

SAFF has the ability to distribute attention effectively by using slots and filtering them using a class prototype as a reference. By combining ideas from metric-based learning and the slot attention mechanism, SAFF can focus the attention on important features removing the non-class relevant features and enabling the creation of meaningful relationships between support and query samples. Fig.~\ref{fig:iterations} shows the iterative refinement process of slot attention across multiple iterations and illustrates the proposed attention per slot and which of them passes the filter and contributes to the final mask.

\section{Methodology} \label{sec:methodology}
In this section, we first define the problem of few-shot learning and present the whole pipeline. Then, we explain in detail the feature filtering mechanism using slot attention. Finally, we describe the classification method based on similarities between query and support patch-filtered embeddings.

\subsection{Problem Definition}

The few-shot image classification problem is commonly defined as an \textit{N}-way \textit{K}-shot task, where \textit{N} denotes the number of classes, and \textit{K} represents the number of labeled examples available per class. This setup includes a support set \textit{S}, which indicates \textit{K} labeled samples for each of the \textit{N} classes, and a query set \textit{Q}, which consists of unlabeled images that must be assigned to one of these \textit{N} classes using the information provided in \textit{S}. 

The support and query set consist of the training set, validation set, and test set, where the test set is fully unseen during the training and validation phases. The objective is to train a model using the training classes so that it can effectively generalize to novel test classes by leveraging the support set. The query set is then used to evaluate the model's classification performance. However, establishing a reliable correspondence between query and support images is challenging due to the limited number of labeled examples per class. The \textit{K}-shot forces models to learn from minimal data, which often results in reduced variability and diversity in the learned features.

\subsection{SAFF Architecture Overview}

\begin{figure*}[t!]
\centering
\includegraphics[width=.925\textwidth, ]{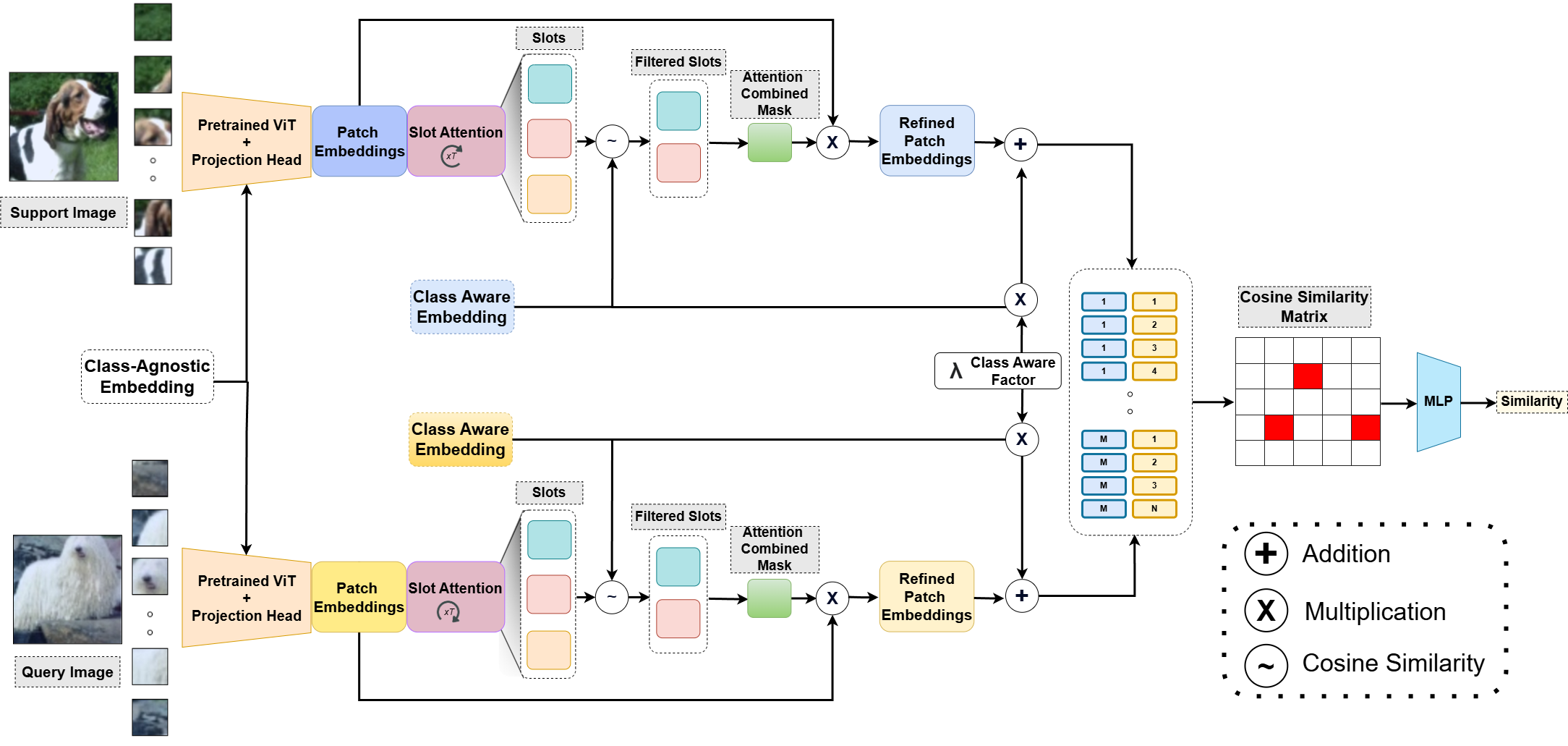}
\caption{Illustration of the SAFF pipeline. First, the ViT-S/16 captures both support and query image information from the input image, generating embeddings for each patch and also feeding the class-agnostic embedding learning the class representation. In the second part, SAFF refines the patch embeddings by using slot attention which is in charge of capturing different attentions and then filtering out the slots that are not similar to the class-aware embedding by removing them. With the selected slots, an attention mask is generated which is then multiplied by the patch embeddings giving weights. Finally, the class-aware embeddings are additionally added with a class-aware factor and the similarity between the query and support refined patches is computed to get the final score.}
\label{fig:pipeline}
\end{figure*}

The pipeline of our method can be seen in Fig.~\ref{fig:pipeline}. Our method works as follows: first, the input images are processed through a pre-trained Vision Transformer (ViT) backbone used as a feature extractor. The ViT splits each image of 224x224 into patches of 16x16 pixels, embeds them through linear projection and gets a class representation capturing global image information. This process generates a sequence of patch $embeddings \in \mathbb{R}^{P\times D}$ where $P=196$ is the number of patches and $D=384$ is the embedding dimension. Then, the patch embeddings are processed by the slot attention module. First, the module is initialized with the class token learned in the previous step to force a better refinement around these features.

Then, through an iterative process of attention, the slots learn to identify and extract patterns by refining the patch embeddings. Afterward, the method discards those attention obtained by slot attention that are furthest away from the class token by computing the similarities between the refined slots and the class token. In this way, an attention-combined mask is created from the average of those slots and then multiplied by the patch embeddings to refine it. Then, we sum the refined embeddings with the class token.

Finally, we compute a dense similarity matrix between the refined patch embeddings of query and support images. This matrix is processed through a Multi-Layer Perceptron (MLP)  that converts the similarities to scores.

\subsection{Slot Attention-Based Feature Filtering}

We integrate a slot attention mechanism to discriminate and underweight non-relevant patches. Through a similarity mechanism, we create a weighted mask with the class-aware slots which is multiplied by the embedding patches resulting in a representative class refined patches. 

In the following subsections, we first detail how the slot attention mechanism is initialized. Next, we examine SAFF which filters out the non-relevant slots and generates the refined embeddings. Finally, we discuss how the refined embeddings are utilized to get the final classification.

\subsubsection{Initialization and Iterative Refinement}

Normally, random initialization is used for slot attention \cite{locatello2020objectcentriclearningslotattention}. In contrast, we use the class token as a seed to specialize the slots to detect attention from patch embeddings based on the class token. Then, the slot attention mechanism refines these initial slots through multiple iterations. In every iteration, the slots compete for the attention of a set of patch-embedding slot attention mechanisms. This iterative process allows slots to progressively focus on different patches with the class token influence.

\subsubsection{Class-Aware Slots Feature Filtering}

After obtaining the refined slots $S_R \in \mathbb{R}^{B \times N \times D}$ where $B$ is the batch size, $N$ the number of defined slots and $D=384$ embedding dimension, we compute their similarity with the class token $C \in \mathbb{R}^{B \times 1 \times D}$. First, we normalize refined slots and the class token using L2 normalization:

\begin{equation}
\hat{\text{S}}_R = \frac{\text{S}_R}{\|\text{S}_R\|_2}, \quad \hat{\text{C}} = \frac{\text{C}}{\|\text{C}\|_2}
\end{equation}

Next, we compute the cosine similarity $d(.,.)$  between $\hat{{\text{S}}_R}$ and $\hat{\text{C}}$ as follows:

\begin{equation}
\text{similarity} = \text{d}(\hat{\text{S}}_R , \hat{\text{C}})
\end{equation}

Then, we apply min-max normalization to scale the cosine similarity score a range between \([0,1]\):
\begin{equation}
\text{similarity}_{\text{norm}} = \frac{\text{similarity} - \text{similarity}_{\min}}{\text{similarity}_{\max} - \text{similarity}_{\min}}
\end{equation}

We create a binary mask for slots with the similarity above mean:
\begin{equation}
    \text{M} = \text{similarity}_{norm} > \text{0.5}
\end{equation}

\begin{equation}
    \text{A}_{masked} = \text{A} \odot \text{M}
\end{equation}

And then computing their average with the filtered number of slots $\text{N}_M$:
\begin{equation}
    \text{A}_{combined} = \frac{\sum_{i=1}^{N} \text{A}_{masked}^i}{\text{N}_M}
\end{equation}

Finally, SAFF weighs the embeddings using the combined attention:
\begin{equation}
    \text{embeddings}_{weighted} = \text{embeddings} \cdot \text{A}_{combined}
\end{equation}

where A represents the attention maps, M is the binary mask, $\odot$ denotes element-wise multiplication and \text{N} is the number of slots. This process ensures that the final weighted embeddings focus on the most class-relevant regions by combining the attention patterns from the most relevant slots.

The threshold of 0.5 provides a good balance between feature selection and information preservation which is particularly important in few-shot scenarios where relevant information can be crucial for correct classification. Also, the results show that a weighted mask instead of a binary mask where only a few patches are filtered allows a smoother transition between relevant and irrelevant features, which is especially important when working with a few samples. In addition, binarization can be too aggressive by completely discarding information that could be useful in a broader context.

Once the embeddings are weighted, we proceed with the linear class-aware addition:
\begin{equation}
    \text{F} = \text{embeddings}_{weighted} + \lambda \cdot \text{classtoken}
\end{equation}

where $\lambda$ is a parameter that controls the influence of the class token. That means we are forcing the patches to have a class-relevant representation. We set $\lambda=2$ following the strategy strategy proposed in \cite{Hao_2023_ICCV}.

\subsection{Classification}
After obtaining the filtered features, we calculate similarities between query and support samples through a dense score matrix:
\begin{equation}
    \text{S}_{ij} = \text{d}(\text{F}_{support}, \text{F}_{query})
\end{equation}

Subsequently, we flatten the similarity matrix $S$ and process it through an MLP to obtain the final classification scores:
\begin{equation}
    \text{scores}_{ij} = \text{MLP}(\text{Flatten}(\text{S}_{ij}))
\end{equation}
where $S_{ij}$ represents the similarity matrix between query image $i$ and support image $j$.

For N-way K-shot classification, we aggregate the similarity scores across the $K$ support samples for each class. For query image $i$ and class $n$, the aggregated score (s) is:
\begin{equation}
   \text{s}_{ni} = \sum_{k=1}^K \text{scores}_{nik}
\end{equation}
where $scores_{nik}$ represents the similarity between query image $i$ and the $k$-th support image of class $n$.

Finally, we compute the classification probabilities (p) using the softmax function:
\begin{equation}
   \text{p}_{ni} = \frac{\exp(s_{ni})}{\sum_{m=1}^N \exp(s_{mi})}
\end{equation}

The model is trained by minimizing the cross-entropy loss over all query images.

\section{Experiments}\label{sec:experiments}

In this section, we provide an introduction to the datasets used and the implementation pipeline. We then discuss the performance of our approach, compare it to the state-of-the-art methods, and provide an extensive analysis of the results obtained.

\subsection{Datasets}
We use four few-shot learning public datasets to validate our proposed method SAFF: 1) {\bf CIFAR-FS} \cite{CIFAR-FS} a subset of CIFAR-100, consists of 100 classes with 600 images per class (32$\times$32 pixels) split into 64 training, 16 validation and 20 testing; 2) {\bf FC100} \cite{FC100} a subset of CIFAR-100, consists of 100 classes with 600 images per class split into 60 training (12 superclasses), 16 validation (4 superclasses) and 20 testing (4 superclasses); 3) {\bf miniImageNet} \cite{MatchingNetwork} a subset of ImageNet, consists of 100 classes with 600 images per class (84$\times$84 pixels) split into 64 training, 16 validation and 20 testing; 4) {\bf tieredImageNet} \cite{tieredImageNet} a subset of ImageNet, consisting of  608 classes split into 350 training (20 superclasses), 97 validation (6 superclasses) and 160 testing (8 superclasses).

Consistent with previous approaches \cite{Hao_2023_ICCV, datasets_method}, we split the dataset into training, validation and test sets. The label spaces are non-overlapping, ensuring that classes present in the training set do not appear in the validation or test sets.

\begin{table*}[!htbp]
\centering
\caption{Comparison of SAFF and CPEA accuracy results on CIFAR-FS, FC100, miniImageNet and tieredImageNet datasets. The Median, Mean and Std. Dev. of the results obtained by fixing three different seeds in the training of each model are shown.}
\label{tab:comparison_seeds}
\begin{tabular}{lcccccccccc}
\toprule
\multirow{2}*{Dataset} & \multirow{2}*{Shot} & \multicolumn{3}{c}{SAFF} & \multicolumn{3}{c}{CPEA\cite{Hao_2023_ICCV}} \\
\cmidrule(lr){3-5} \cmidrule(lr){6-8}
& & Median & Mean & Std. Dev. & Median & Mean & Std. Dev. \\
\midrule
CIFAR-FS & 1 & \textbf{78.48} & 78.50 & 0.30 & 78.32 & 78.39 & 0.23 \\
CIFAR-FS & 5 & \textbf{90.30} & 90.26 & 0.08 & 89.50 & 89.49 & 0.16 \\
FC100 & 1 & \textbf{47.17} & 47.17 & 0.12 & 46.70 & 46.81 & 0.53 \\
FC100 & 5 & \textbf{66.00} & 66.22 & 0.51 & 65.70 & 65.57 & 0.35 \\

miniImageNet & 1 & \textbf{71.51} & 71.48 & 0.17 & 71.34 & 71.32 & 0.45 \\
miniImageNet & 5 & \textbf{87.19} & 87.20 & 0.06 & 86.73 & 86.70 & 0.15 \\
tieredImageNet & 1 & \textbf{74.73} & 74.70 & 0.30 & 74.58 & 74.51 & 0.57 \\
tieredImageNet & 5 & \textbf{88.97} & 89.00 & 0.09 & 88.71 & 88.71 & 0.35 \\

\bottomrule
\end{tabular}
\end{table*}

\subsection{Experimental Settings}

\begin{table*}[!htbp]
\centering
\caption{Comparison with state-of-the-art methods on 5-way 1-shot and 5-way 5-show with 95 \% confidence intervals on CIFAR-FS and FC100}
\label{tab:comparison_cifar}
\resizebox{\textwidth}{!}{
\begin{tabular}{lccccccc}
\toprule
\multirow{2}*{Model} & \multirow{2}*{Backbone} & \multirow{2}*{$\approx$ \# Params} & \multicolumn{2}{c}{CIFAR-FS} & \multicolumn{2}{c}{FC100} \\
\cmidrule(lr){4-5} \cmidrule(lr){6-7}
& & & 1-shot & 5-shot & 1-shot & 5-shot \\
\midrule
ProtoNet \cite{NIPS2017_cb8da676} & ResNet-12 & 12.4 M & - & - & 41.54$\pm$0.76 & 57.08$\pm$0.76 \\
RENet \cite{RENet} & ResNet-12 & 12.4 M & 74.51$\pm$0.46 & 86.60$\pm$0.32 & - & - \\
TPMN \cite{TPMN} & ResNet-12 & 12.4 M & 75.50$\pm$0.90 & 87.20$\pm$0.60 & 46.93$\pm$0.71 & 63.26$\pm$0.74 \\
PSST \cite{PSST} & WRN-28-10 & 36.5 M & 77.02$\pm$0.38 & 88.45$\pm$0.35 & - & - \\
Meta-QDA \cite{MetaQDA} & WRN-28-10 & 36.5 M & 75.83$\pm$0.88 & 88.79$\pm$0.75 & - & - \\
FewTURE \cite{FewTURE} & ViT-S/16 & 22 M & 76.10$\pm$0.88 & 86.14$\pm$0.64 & 46.20$\pm$0.79 & 63.14$\pm$0.73 \\
\midrule
CPEA* & ViT-S/16 & 22 M & 78.32$\pm$0.65 & 89.50$\pm$0.42 & 46.70$\pm$0.58 & 65.70$\pm$0.57 \\
\textbf{SAFF (ours)} & ViT-S/16 & 22 M & \textbf{78.48$\pm$0.67} & \textbf{90.30$\pm$0.41} & \textbf{47.17$\pm$0.59} & \textbf{66.00$\pm$0.57} \\
\bottomrule
\multicolumn{7}{l}{\textbf{* The median results obtained in each experiment for CPEA \cite{Hao_2023_ICCV}.}}
\end{tabular}
}
\end{table*}

\begin{table*}[!htbp]
\centering
\caption{Comparison with state-of-the-art methods on 5-way 1-shot and 5-way 5-show with 95 \% confidence intervals on miniImageNet and tieredImageNet }
\label{tab:comparison}
\resizebox{\textwidth}{!}{
\begin{tabular}{lccccccc}
\toprule
\multirow{2}*{Model} & \multirow{2}*{Backbone} & \multirow{2}*{$\approx$ \# Params} & \multicolumn{2}{c}{miniImageNet} & \multicolumn{2}{c}{tieredImageNet} \\
\cmidrule(lr){4-5} \cmidrule(lr){6-7}
& & & 1-shot & 5-shot & 1-shot & 5-shot \\
\midrule
ProtoNet \cite{NIPS2017_cb8da676} & ResNet-12 & 12.4 M & 62.29$\pm$0.33 & 79.46$\pm$0.48 & 68.25$\pm$0.23 & 84.01$\pm$0.56 \\
FEAT \cite{FEAT} & ResNet-12 & 12.4 M & 66.78$\pm$0.20 & 82.05$\pm$0.14 & 70.80$\pm$0.23 & 84.79$\pm$0.16 \\
InfoPatch \cite{InfoPatch} & ResNet-12 & 12.4 M & 67.67$\pm$0.45 & 82.44$\pm$0.31 & - & - \\
FEAT \cite{FEAT} & WRN-28-10 & 36.5 M & 65.10$\pm$0.20 & 81.11$\pm$0.14 & 70.41$\pm$0.23 & 84.38$\pm$0.16 \\
OM \cite{OM} & WRN-28-10 & 36.5 M & 66.78$\pm$0.30 & 85.29$\pm$0.41 & 71.54$\pm$0.29 & 87.79$\pm$0.46 \\
FewTURE \cite{FewTURE} & ViT-S/16 & 22 M & 68.02$\pm$0.88 & 84.51$\pm$0.53 & 72.96$\pm$0.92 & 86.43$\pm$0.67 \\
\midrule
CPEA* & ViT-S/16 & 22 M & 71.34$\pm$0.64 & 86.73$\pm$0.38 & 74.58$\pm$0.72 & 88.71$\pm$0.46 \\
\textbf{SAFF} (ours) & ViT-S/16 & 22 M & \textbf{71.51$\pm$0.63} & \textbf{87.19$\pm$0.36} & \textbf{74.73$\pm$0.74} & \textbf{88.97$\pm$0.43} \\
\bottomrule
\multicolumn{7}{l}{\textbf{* The median results obtained in each experiment for CPEA \cite{Hao_2023_ICCV}.}}
\end{tabular}
}
\end{table*}

\begin{table}[!htbp]
\centering
\begin{tabular}{lc}
\toprule
\textbf{Attention Mechanism} & \textbf{Accuracy (\%)}\\
\midrule
Dot-Product Attention & 90.00±0.40 \\
\midrule
Cross Attention Query-Support & 89.97±0.41\\
\midrule
Cross Attention Patch-Token & 90.03±0.41\\
\midrule
Slot Attention & \textbf{90.30±0.41}\\

\bottomrule
\end{tabular}
\caption{Comparison with different attention modules on 5-way 5-show with 95 \% confidence intervals on CIFAR-FS.}
\label{tab:attention_mechanism}
\end{table}

We use a ViT-B/16 \cite{VIT16} pretrained using the \cite{tokenizer} strategy. We followed the choice of the backbone and hyperparameters settings established by \cite{Hao_2023_ICCV} in order to compare correctly and fairly. The slot attention module is configured with 5 slots, embeddings dimension 384 as backbone output, 5 iterations and initialized with class token embedding.
      
We report the test accuracy of image recognition as a performance metric. To ensure the robustness of our results and fair comparison with the best state-of-the-art method, we have used three different seeds in our experiments and reported the median, mean and standard deviation (Std. Dev.) in Table \ref{tab:comparison_seeds} and the median results in the rest of the tables.

Our experimental evaluation follows the standard few-shot learning protocol, conducting tests on 5-way 1-shot and 5-way 5-shot scenarios. For each dataset, we evaluate the model's performance over 1,000 randomly generated test episodes, where each episode contains 15 query samples per class, and give the mean classification accuracy.

\subsection{Performance Comparison to the SoTA}

\begin{table}[htbp]
\centering
\begin{tabular}{lccc}
\toprule
\multirow{2}{*}{\textbf{Dataset}} & \multicolumn{3}{c}{\textbf{SAFF vs. CPEA}} \\
\cmidrule(lr){2-4}
 & $\chi^2$ & p-value & Significance \\
\midrule
CIFAR-FS     & 2.02  & 0.16 & \xmark \\
TieredImageNet & 75.97 & 2.87$\times$10$^{-18}$ & \cmark \\
MiniImageNet   & 36.59 & 4.17$\times$10$^{-5}$ & \cmark \\
CIFAR100     & 43.97 & 3.32$\times$10$^{-11}$ & \cmark \\
\bottomrule
\end{tabular}
\caption{McNemar's test evaluating the statistical significance between SAFF and CPEA for 5-way 1-shot classification. Significant = $p-value \leq 0.001$, Not Significant = $p-value > 0.001$.}
\label{tab:mcnemar_results}

\end{table}

The results across different seeds demonstrate that SAFF surpasses the state-of-the-art methods and shows more stability minimizing the variance. As shown in Table \ref{tab:comparison_cifar}, on CIFAR-FS, the accuracy reaches \textbf{90.30\%} in the 5-shot setting and \textbf{78.48\%} in the 1-shot setting, improving \textbf{0.8\%} and \textbf{0.16\%} respectively compared to CPEA. For FC100, SAFF achieves \textbf{66.00\%} accuracy in the 5-shot setting and \textbf{47.17\%} in the 1-shot setting representing a \textbf{0.30\%} and \textbf{0.47\%}  improvement over CPEA. As shown in Table \ref{tab:comparison}, in miniImageNet, our model attains \textbf{87.19\%} accuracy for 5-shot tasks, surpassing CPEA by \textbf{0.46\%} and \textbf{71.51\%} accuracy for 1-shot tasks, surpassing CPEA by \textbf{0.17\%}. Finally, on tieredImageNet, SAFF achieves \textbf{88.97\%} in the 5-shot settings and \textbf{74.73\%} in the 1-shot settings representing a \textbf{0.26\%} and \textbf{0.15\%}  improvement over CPEA.

To validate the statistical significance of our improvements, we conducted McNemar's test, based on \cite{nemar}, on the model predictions as shown in Table \ref{tab:mcnemar_results}. The results reveal that SAFF achieves statistically significant improvements (p $\leq$ 0.01) over CPEA on three out of four datasets. 

These results consistently show that SAFF not only achieves superior performance, but also maintains remarkable stability across different random seeds as shown in Table \ref{tab:comparison_seeds} where we compare only with CPEA as other methods fall significantly short of our results. This is particularly evident in the 5-shot scenarios where it consistently outperforms previous state-of-the-art methods.

\subsection{Analysis of Slot-Attention Hyperparameters}

The two main components of the slot-attention mechanism are the number of slots and the number of iterations. In \cite{locatello2020objectcentriclearningslotattention}, the influence of these parameters on model performance in a semantic segmentation approach is analyzed. Specifically, a larger number of slots tends to not necessarily improve performance, with optimal results achieved at 5 slots and performance decreasing with 10 slots.  On the other hand, a larger number of iterations produces minimal impact on the accuracy, with the results remaining stable across 3, 5 and 10 iterations. In this subsection, we analyze the influence of these slot attention components for the FSL problem considering the proposed method. The experimental results in Table~\ref{tab:iterations} show that the optimal configuration uses 5 slots with 5 iterations achieving \textbf{90.30$\pm$0.41} accuracy on CIFAR-FS and \textbf{87.19$\pm$0.36} accuracy on miniImageNet. This suggests that moderate values for both parameters provide an optimal balance between capacity and computational efficiency, as increasing either parameter does not yield significant improvements.

\subsection{Evaluation of Attention Mechanisms and Filtering Strategies}

Based on Table~\ref{tab:attention_mechanism}, the Slot Attention mechanism demonstrates the highest accuracy at \textbf{90.30$\pm$0.41} suggesting its superior capability to extract and focus on relevant features. The improvement comes from its ability to decompose complex visual scenes into object-centric representations through iterative refinement. This decomposition process creates more discriminative representations that benefit few-shot classification tasks, where limited examples are available for learning.

Following our implementation of slot attention, we apply a combined attention mask to reduce the weight of some patches that may not be relevant. Two strategies are considered for filtering patches: a) \textit{binary}, in which patches related to an activation lower than 0.5 are completely canceled (set to 0) and b) \textit{weighted}, in which the activation is considered to reduce the importance of less relevant patches.

Table~\ref{tab:mask_comparison} shows the results of both binary and weighted masking strategies on various datasets for 5-shot learning.
Results show that weighted masking improves feature discrimination over binary decisions. Using weighted masking, SAFF can preserve contextual information while increasing the weight on the most representative patches.

\begin{table}[th]
\centering
\begin{tabular}{llcc}
\toprule
\textbf{Iterations} &\textbf{\# Slots} &\textbf{CIFAR-FS} &\textbf{miniImageNet} \\
\midrule
3 & 3 & 89.40±0.46 & 86.56±0.39 \\
5 & 3& 89.46±0.46 & 86.65±0.38 \\
10 & 3 & 89.46±0.46 & 86.61±0.39\\
\midrule
3 & 5& 90.28±0.41 & {87.10±0.37} \\
5 & 5& \textbf{90.30±0.41} & \textbf{87.19±0.36}\\
10 & 5 & 90.29±0.40 & {87.16±0.36}\\
\midrule
3 & 10 & 89.61±0.45 & 86.95±0.40\\
5 & 10& 89.63±0.45 & 87.09±0.39\\
10 & 10 & 89.52±0.45& 87.01±0.40\\

\bottomrule
\end{tabular}
\caption{Number of Slots/Iterations vs Accuracy (\%) for 5-shot learning on CIFAR-FS and miniImageNet.}
\label{tab:iterations}
\end{table}

\begin{table}[th]
\centering
\begin{tabular}{lccc}  
\toprule
\textbf{Mask} &\textbf{CIFAR-FS} &\textbf{FC100} &\textbf{miniImageNet}\\
\midrule
Binary & 90.20±0.45 & 65.87±0.56 & 87.04±0.42 \\
Weighted & \textbf{90.30±0.41} & \textbf{66.00±0.57} & \textbf{87.19±0.36} \\
\bottomrule
\end{tabular}
\caption{Accuracy (\%) vs Binary/Weighted mask for 5-shot learning using 5 slots.}
\label{tab:mask_comparison}
\end{table}

\section{Conclusions}
\label{sec:conclusions}

The proposed SAFF approach demonstrates superior performance across multiple benchmark datasets, particularly excelling in scenarios with limited training examples. Through comprehensive experimental evaluation, SAFF improves state-of-the-art results on CIFAR-FS, FC100, miniImageNet and tieredImageNet datasets.
%
%
The consistent performance improvements across various benchmarks can be attributed to two
key aspects. First, the slot-based feature fusion mechanism effectively captures and combines discriminative features from both support and query samples, enhancing the model's ability to learn from limited examples. Second, the weighted masking strategy enables more precise feature selection by providing continuous attention values, leading to smoother transitions between relevant and irrelevant features compared to traditional binary approaches. 

However, the slot attention mechanism is highly dependent on the number of slots and iterations defined. Moreover, slot attention may have difficulty capturing complex attention leading to degraded performance in environments where distinguishing the main class from noise is challenging.
To address these challenges, 
in future work, we will analyze the definition of
a dedicated slot for each class available in the dataset enabling each slot to specialize in capturing specific attention to its respective class. 
This can help to mitigate the need to predefine the number of slots. 

\section*{Acknowledgements} This work has been partially supported by the Horizon EU project MUSAE (No. \museNo), 2021-SGR-01094 (AGAUR), Icrea Academia'2022 (Generalitat de Catalunya), Robo STEAM (2022-1-BG01-KA220-VET-000089434, Erasmus+ EU), DeepSense (ACE053/22/000029, ACCIÓ), DeepFoodVol (AEI-MICINN, \DFVolNo), PID2022-141566NB-I00 (AEI-MICINN), and Beatriu de Pinós Programme and the Ministry of Research and Universities of the Government of Catalonia (2022 BP 00257).

\begingroup
\bibliographystyle{ieeenat_fullname}
\bibliography{main}
\endgroup
\end{document}